\documentclass[conference]{IEEEtran}
\usepackage{graphicx}
\usepackage{times}
\usepackage{tabularx}
\usepackage{xspace}
\usepackage{comment}
\usepackage{todonotes}

\usepackage{xcolor}

\usepackage[numbers]{natbib}
\usepackage{multicol}
\usepackage[bookmarks=true]{hyperref}

\newcommand{\etal}{et al.\@\xspace}
\newcommand{\ie}{i.e.\@\xspace}
\newcommand{\eg}{e.g.\@\xspace}
\newcommand{\etc}{etc.\@\xspace}

\newcommand\blfootnote[1]{%
  \begingroup
  \renewcommand\thefootnote{}\footnotetext{#1}%
  \addtocounter{footnote}{-1}%
  \endgroup
}

\begin{document}

\title{Demonstrating Mobile Manipulation in the Wild:\newline A Metrics-Driven Approach}


\author{\authorblockN{Max Bajracharya, James Borders, Richard Cheng, Dan Helmick, Lukas Kaul, Dan Kruse, \\ John Leichty, Jeremy Ma, Carolyn Matl, Frank Michel, Chavdar Papazov, \\ Josh Petersen, Krishna Shankar, Mark Tjersland\authorrefmark{1}}
\authorblockA{Toyota Research Institute (TRI)\\
Los Altos, California 94022\\
Email: firstname.lastname@tri.global}}

\maketitle


\blfootnote{\authorrefmark{1}All authors contributed equally and are listed in alphabetical order.}

\begin{abstract}
We present our general-purpose mobile manipulation system consisting of a custom robot platform and key algorithms spanning perception and planning. 
To extensively test the system in the wild and benchmark its performance, we choose a grocery shopping scenario in an actual, unmodified grocery store. 
We derive key performance metrics from detailed robot log data collected during six week-long field tests, spread across 18 months.
These objective metrics, gained from complex yet repeatable tests, drive the direction of our research efforts and let us continuously improve our system’s performance. 
We find that thorough end-to-end system-level testing of a complex mobile manipulation system can serve as a reality-check for state-of-the-art methods in robotics. 
This effectively grounds robotics research efforts in real world needs and challenges, which we deem highly useful for the advancement of the field. 
To this end, we share our key insights and takeaways to inspire and accelerate similar system-level research projects.
\end{abstract}

\IEEEpeerreviewmaketitle
\section{Introduction} \label{sec:intro}
The ultimate goal of mobile manipulation research is to enable capable mobile manipulation robots to perform human-level tasks in diverse and complex environments. However, despite the vast robotics research in the past decades, autonomous mobile manipulators still struggle to operate in the real world, outside of specifically designed or modified environments. This should come as no surprise - mobile manipulation in human environments is an extremely challenging endeavor, requiring that robots:
\begin{itemize}
    \item Accurately perceive their environment to estimate their own location, detect obstacles, landmarks, and objects of interest.
    \item Quickly generate collision-free motion plans to/from objects and locations of interest while avoiding obstacles.
    \item Manipulate a wide variety of unseen objects in ever-changing environments.
\end{itemize}

Each of the aforementioned tasks constitutes a difficult research problem on its own. This is why much of today's robotics research is focused on individual challenges in the robotic pipeline. While this has led to significant progress in developing individual aspects of the mobile manipulation problem (\eg, motion planning, scene segmentation, grasping, \etc), relatively little work has been done that combines these capabilities and evaluates them in real-world environments. This bears the risk of leaving important challenges---crucial for real-world deployment---unnoticed, while arguably less critical aspects are overemphasized. For example, motion planning research continues to push for more optimal trajectories in minimal time, while often settling for 90-95\% success rates. However, at a systems level, the effect of a single percentage drop in success rate far outweighs even a significant drop in path optimality or computation time.
Therefore, we believe that identifying and tackling fundamental problems that stand in the way of more widespread, real-world robot deployments by studying system-level performance in the field is an important contribution towards advancing the field of robotics.

\begin{figure}[t]
\includegraphics[width=0.45\textwidth]{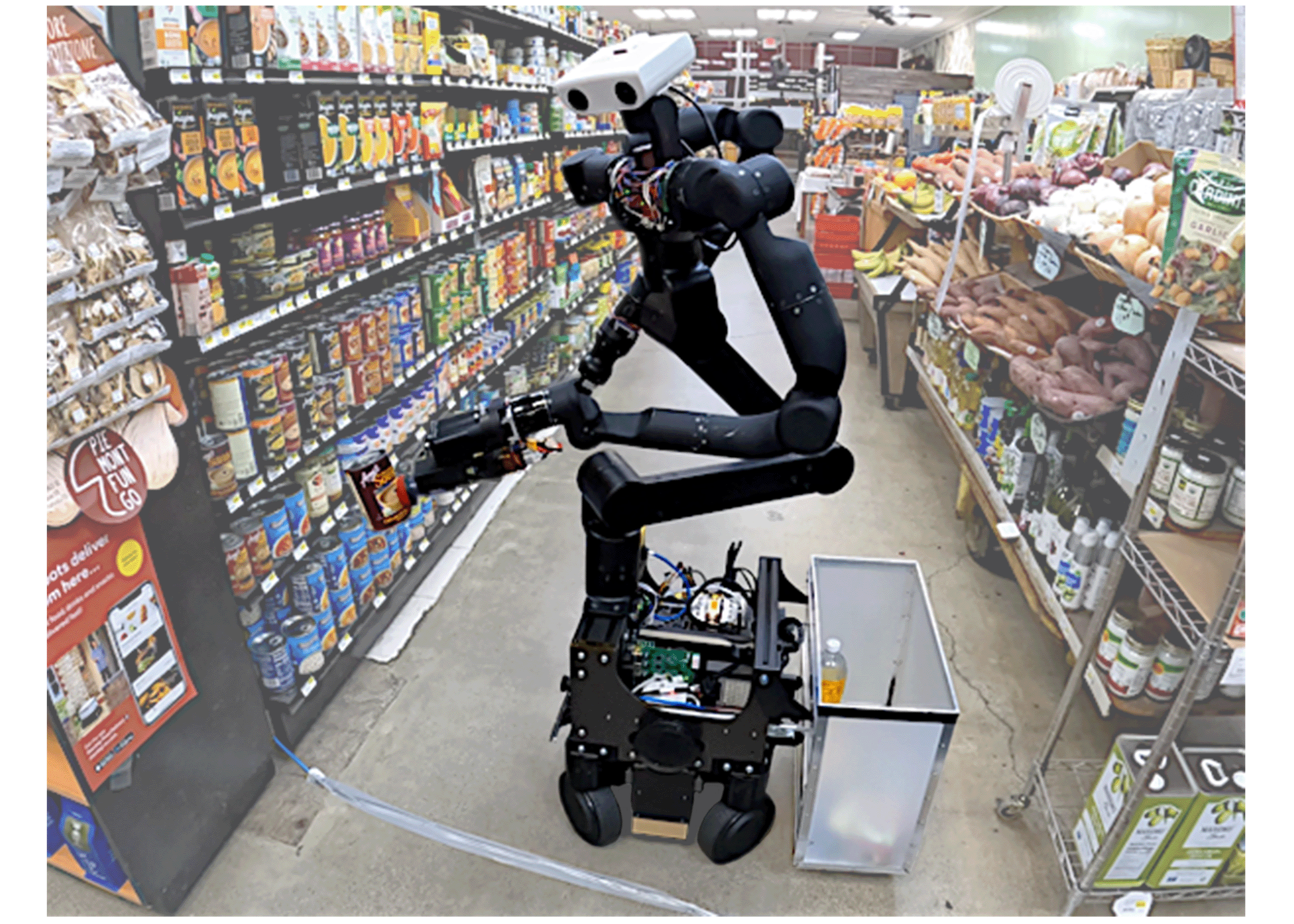}
\caption{TTT operating in a real, unmodified grocery store.}
\label{fig:robot}
\vspace{-0.55cm}
\end{figure}

In order to tackle the aforementioned difficulties and to push the development of capable mobile manipulation robots, we created a challenge task: a robot should go into a real grocery store, pick 20 unique items off the shelves from a randomly generated ``shopping list'' (out of $\sim$1000 potential items), place them in a basket, and bring them to its starting position.
Figure~\ref{fig:robot} shows our robot in the grocery store we use for testing. Note that it is a real store rather than a lab replica, and that we do not modify it in any way for our tests. In addition to forcing us to address all the previously mentioned challenges (\eg, perception, planning, grasping), task performance is easily quantifiable, allowing objective measurement of performance and progress. 

This paper describes our robotic platform TTT, its custom hardware design, and algorithms that we developed to tackle the shopping challenge task.
We focus on the entire mobile manipulation system rather than discussing any single subsystem or algorithm in detail. Most importantly, by developing and testing the system end-to-end over an extended period of time without abstracting away any components, we glean valuable insights into what important, unsolved challenges are for deploying mobile manipulators in real, semi-structured environments.

Section~\ref{sec:related} briefly discusses related work. 
Section~\ref{sec:sys_overview} provides a system overview and describes the system architecture. 
Sections~\ref{sec:hw}, \ref{sec:perception} and \ref{sec:plan} describe the hardware, perception and behavior systems, respectively. 
Section~\ref{sec:eval} evaluates the robot's performance on the grocery shopping challenge task. 
Section~\ref{sec:conclusion} concludes by discussing important lessons learned and avenues for future research.
This paper is accompanied by a video that shows our robot in operation during the most recent (sixth) field test at the grocery store.

\section{Related Work} \label{sec:related}
Robotic mobile manipulation is a highly active, multi-faceted area of research.
Interest in the field is driven by numerous potential applications and amplified by international competitions such as DARPA’s robotics challenge \cite{Guizzo-2015-Spectrum}, RoboCup@Home \cite{wisspeintner-2009-robocup}, the Amazon Picking Challenge \cite{eppner2016lessons} and RoboCup@Work \cite{Kraetzschmar-2015-Robo}, each focusing on different challenges and performance criteria.

A substantial number of capable mobile manipulation platforms have resulted from past and current research projects.
Systems that were influential in the class of wheeled dual-arm mobile manipulators like the one we are presenting include Willow Garage's PR2 \cite{Meeussen-2010-ICRA}, CMU's Herb2.0 \cite{herb2_journal}, DLR's Rollin' Justin \cite{borst-2009-rollin}, JPL's RoboSimian \cite{robosimian}, KIT's ARMAR-6 \cite{asfour-2019-armar} and IIT's CENTAURO \cite{centauro}.
An overview over wheeled mobile manipulation systems and the challenges involved is provided in \cite{Thakar-2023-MaR} and \cite{Sereinig-2020}.
Particularly related to the work we present are studies that involve the testing of mobile manipulation systems in semi-structured environments outside the lab.
D{\"o}mel \etal \cite{domel-2017-toward} used a wheeled, single arm mobile manipulation system to fulfill fetch and carry tasks in a factory environment, similar to our shopping scenario. They conducted a daylong evaluation test to asses their system's performance.  
\v{S}tibinger \etal \cite{vstibinger2021mobile} used a morphologically similar system in an outdoor competition to pick up and place simulated construction materials.

To the best of our knowledge, our study is the first to describe an approach for improving a fully autonomous dual-arm mobile manipulation system based on rigorous, repeated testing in an unmodified, real-world environment over an extended period of time.

\section{System Overview}
\label{sec:sys_overview}
\begin{figure}[t]
\includegraphics[width=0.48\textwidth]{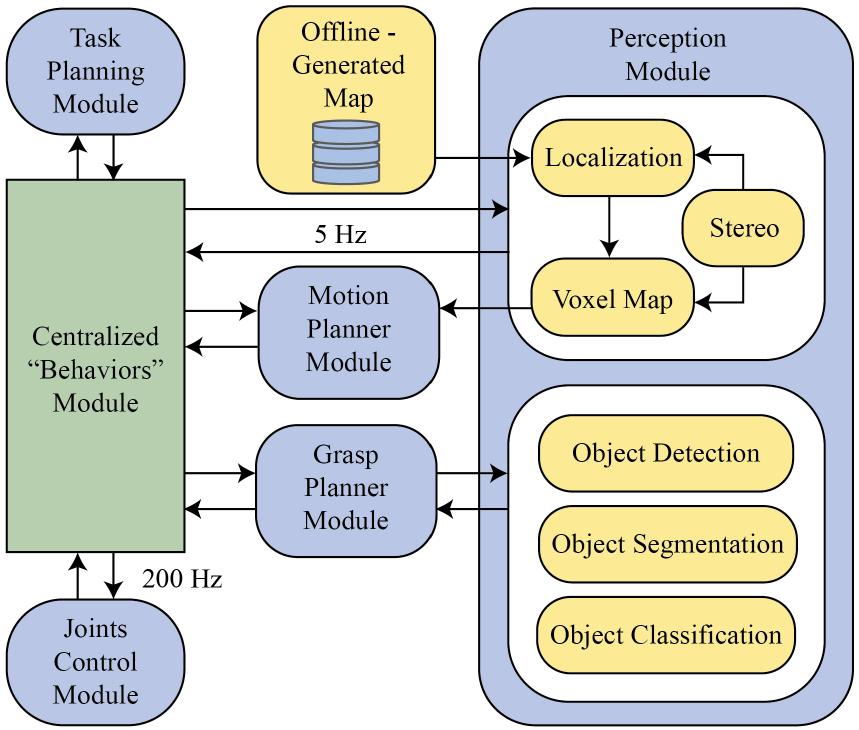}
\caption{Overview of our interconnected, modular software system designed for mobile manipulation tasks.}
\label{fig:sys_overview}
\vspace{-0.35cm}
\end{figure}
Figure~\ref{fig:sys_overview} provides an overview of the software system. The main system capabilities are encapsulated within \textit{modules} running on board at different rates, communicating through a custom interprocess communication (IPC) framework.

At the highest level, a task planning module implements a hierarchical finite state machine (FSM), which keeps track of the robot's state and determines what high-level action it should take next (\eg, look for item, grasp item, localize, \etc). 
At the core of the system is a centralized \textit{behavior} module. Once the task planning module decides which action to take,  the behavior module queries other perception/planning modules to aggregate all necessary information and send low-level commands to the hardware at 200Hz in order to achieve desired objectives. 
A centralized behavior module that has exclusive control over joints simplifies the system architecture and reduces potential for conflicts between modules.

A visual perception module, running at 5 Hz, keeps the robot localized and aware of obstacles by generating a stereo point cloud and a voxel map for spatial awareness. Once the robot needs to grasp an item, it queries an event-triggered grasp planning module, and whenever the robot needs to plan a motion, it queries an event-triggered motion planning module. In the following sections, we describe these modules (and sub-modules) in more detail. 

While many of these modules are based on pre-existing methods, there was significant benefit for us to implement them from the ground up.
This way we can ensure to have complete insight into and control over the implementation, allowing us to (1) integrate tooling/visualizations to debug issues with minimal overhead, and (2) easily and continously modify/customize/improve our algorithms and implementations based on the results of our testing.

\section{Hardware System} \label{sec:hw}
Our mobile manipulator robot TTT consists of four distinct parts: (1) a pseudo-holonomic 4-wheeled chassis; (2) a 5-DoF torso; (3) two 7-DoF arms; and (4) a perception head on a 2-DoF neck. Nearly all components have been custom-designed from a low level, creating a homogeneous system that minimizes the footprint and upper body mass of the robot while maximizing its manipulation capabilities, particularly its range of motion and payload capacity.

\noindent
\textbf{Actuation:} To maximize the payload capability of the robot, we developed three sizes of rotary actuators that are tightly integrated and have very high specific torques and torque densities. Each actuator has an integrated custom motor controller, torque sensor, and output position encoder. The controllers are capable of communicating via CAN and EtherCAT. We currently prefer using CAN.

\noindent
\textbf{Tools:} At the end of each arm is a tool interface that allows tools to be manually installed and removed. For the grocery shopping task, we use a Robotiq HandE gripper on the right arm and a custom suction gripper on the left arm. In contrast to most industrial applications of suction grippers that primarily perform top grasps, we exclusively use lateral side grasps for the grocery task due to the tight constraints of the shelf. This is a more challenging application for suction grasping due to the gravity-induced moment and shear force acting on the suction cup.

Our custom suction gripper integrates two vacuum generators, a Robotiq EPick with an integrated diaphragm pump capable of producing an $82\%$ vacuum, and a high-speed axial pump able to generate much larger flow rates, albeit at a lower vacuum level. We use the high flow rate when approaching objects to help create initial contact and to hold onto bags that are inherently challenging for suction grasps. If the tool's internal pressure signals a sufficient seal, it switches over to the diaphragm pump, taking advantage of its much stronger vacuum.

\noindent
\textbf{Compute:} The robot has a single central compute system consisting of a standard ATX motherboard with an Intel Core i9-12900K CPU and an NVIDIA A6000 GPU. It has 15 TB of removable U.2 NVMe storage for logging. The use of modular consumer components as opposed to embedded single board computers allows us to take advantage of new hardware as soon as it is released. The robot communicates over Wi-Fi 6 to a consumer mesh network system.

\noindent
\textbf{Sensing:} We exclusively use stereo cameras for visual perception, with a pair of Basler acA2500-60uc color cameras with wide-angle lenses mounted on the pan/tilt head and another front-facing pair mounted to the chassis. Each arm has a Sunrise Instruments M3553E 6-axis force/torque sensor mounted between the wrist and the tool interface.

\noindent
\textbf{Power:} To allow several hours of uninterrupted field testing, the robot has a large amount of energy storage contained in four hot-swappable battery packs. Each pack is made up of three BB-2590 lithium-ion batteries and a backplane adapter, for a total of 3.6 kWh of capacity. The packs plug directly into a custom power board stack in the center of the chassis which monitors and controls the robot power system. 

\section{Perception Modules} \label{sec:perception}
\subsection{Learned Stereo Depth and Voxel Mapper} \label{sec:stereo}
\noindent
The input to the visual perception system consists of color stereo image pairs. A learned stereo algorithm \cite{Shankar-2022-RAL} is used to produce dense and accurate depth from camera pairs in the robot head and chassis. The resulting RGB-D frames are fused into a dynamic 3D voxel map \cite{bajracharya2012real,bajracharya2020mobile}. Each voxel accumulates color as well as first and second order position statistics. 

\subsection{Object Detection, Segmentation, and Classification} \label{sec:object}
Object perception is crucial for both grasping and creating a map that contains the items in the store. It is done in the following way:
\begin{itemize}
\item \textbf{Object Detection:} We run a YOLOv5 detector \cite{Glenn-2020} trained on SKU-110k \cite{Goldman-2019-CVPR} to obtain 2D bounding boxes around all items in an input image.
\item \textbf{Object Segmentation:} We utilize a UNet-based segmentation network \cite{UNet}, trained on a mix of synthetic and real data, in order to obtain a segmentation mask of the item within the bounding box. The item pointcloud is then obtained by overlaying this segmentation mask with the RGB-D image obtained with the learned stereo module described in Section~\ref{sec:stereo} and back-projection to 3D.
\item \textbf{Object Classification:} Given a query image crop using the bounding boxes from the detector, a metric-learning-based approach \cite{Koch-2015-ICML} is used to find the most similar items in a database of images that are scraped from the internet. A second stage Prototypical Network \cite{Snell-NIPS-2017} gives a fine-grained match and is trained with open-set loss \cite{Liu-2020-CVPR} to handle the case when the query item does not exist in the database.
\end{itemize}

\begin{figure}[ht]
\includegraphics[width=0.48\textwidth]{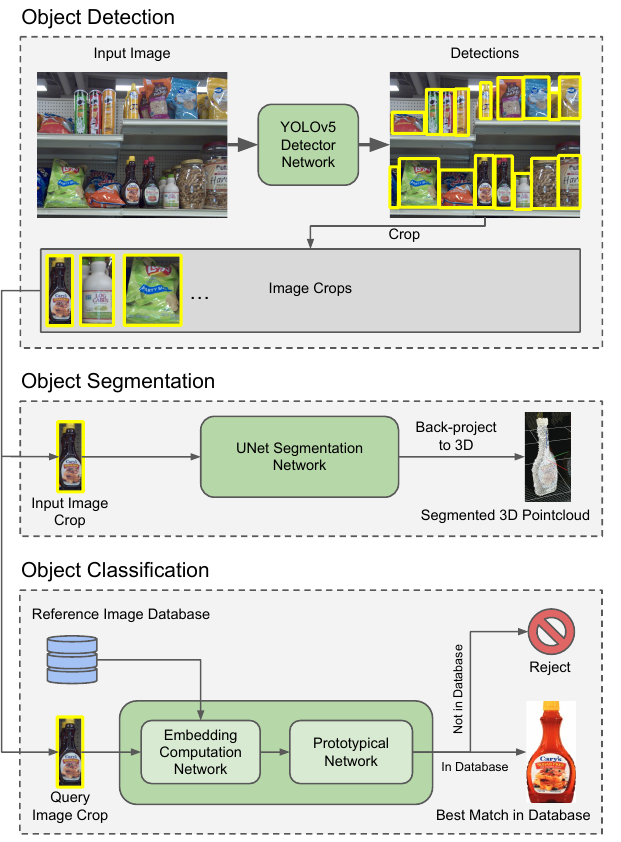}
\caption{The object perception pipeline as outlined in Section~\ref{sec:object}.}
\label{fig:item_perception}
\vspace{-0.4cm}
\end{figure}

This provides us with segmented pointclouds of the items in the scene, along with their classification. The pipeline is shown in Figure~\ref{fig:item_perception}, and it is utilized to map items within the store prior to robot deployment, and also by the robot when picking items from the shelves and placing them in its basket.

\subsection{Mapping and Localization} \label{sec:lm}
To enable our mobile manipulation robot to shop for groceries, it has to know its own pose as well as where items are located in the store. Since the venue is known in advance, there is no need to pursue a SLAM approach. Instead, we treat the two tasks separately: Mapping is done prior to robot deployment, and localization within the generated map is performed by the robot during task executions.

\noindent \textbf{Mapping:} 
The input to the mapping pipeline is a time-ordered stream of stereo images $(L_0, R_0), \ldots, (L_n, R_n)$, acquired with stereo cameras that are moved through the store. We run our learned stereo network on each image pair to compute per-pixel depth. We detect image keypoints in each image with a DoG detector \cite{Lindeberg-1993-IJCV} and store a RootSIFT \cite{Arandjelovic-2012-CVPR} descriptor for every keypoint that has valid depth. The map is then generated via the following four steps:
\begin{figure*}[ht]
\includegraphics[width=\textwidth]{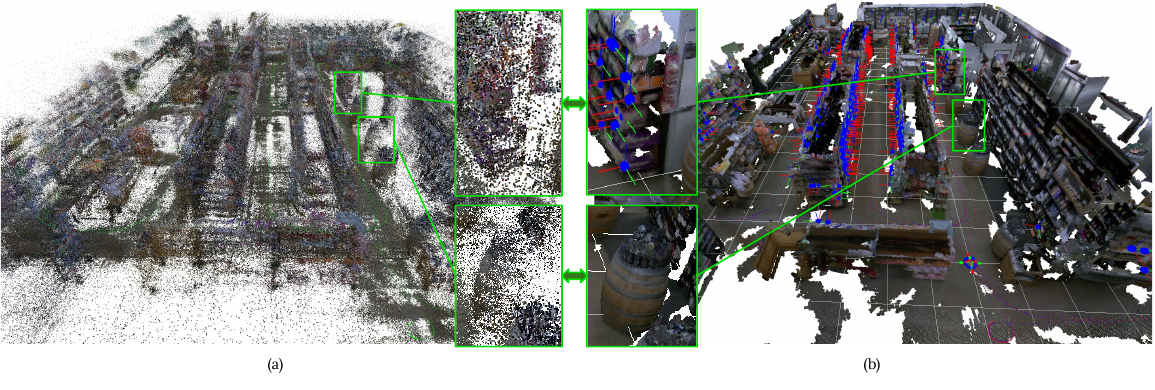}
\caption{(a) Output of a state-of-the-art BA algorithm \cite{Agarwal-2022}. Notice the noise and sparsity of the reconstruction. (b) Our map generated by registering the stereo-based 3D pointclouds using the poses computed by the same BA algorithm. The green rectangles show a side-to-side comparison between the same two regions in the maps and highlight the difference in reconstruction quality. The positions of the mapped store items are rendered as blue spheres and their orientations are indicated by red, blue and green coordinate frames.}
\vspace{-0.4cm}
\label{fig:mapping}
\end{figure*}
\begin{itemize}
\item \textbf{Visual Odometry (VO):} We estimate camera pose deltas between each pair of subsequent left images $(L_{k-1}, L_k)$. First, we match the descriptors in $L_{k-1}$ to their nearest neighbors in $L_k$ and vice versa. We only keep mutual nearest neighbors as valid correspondences. Next, we compute the pose of $L_k$ relative to $L_{k-1}$ by running a PnP solver \cite{Lepetit-2009-IJCV} within a RANSAC loop \cite{Fischler-1981-ACM}.
\item \textbf{Loop detection and closure:} Since VO leads to pose drift, we perform loop detection and correct for the accumulated VO pose error. We train a NetVLAD-based network \cite{Arandjelovic-2016-CVPR} to output a global image descriptor $\mathbf g_k$ for each image $L_k$. To detect a loop closure, we find the 5 nearest neighbors to $\mathbf g_k$ and compute pose deltas between the corresponding images and $L_k$ in the same way as in the VO section above. We accept the pose delta with the highest number of inliers (if that number exceeds 200). Then, we correct for the pose drift using a deformation modeling technique, similar to \cite{ElasticFusion}.
\item \textbf{Bundle Adjustment (BA):} We pass the following input to a state-of-the-art BA algorithm \cite{Agarwal-2022}: (i) the 2D locations of the inlier keypoints, (ii) initial estimates of the 3D coordinates of the physical points we want to reconstruct (computed by averaging the 3D coordinates of corresponding keypoints) and (iii) the camera poses computed by VO and loop closures. The output is a set of refined 3D point coordinates and camera poses. However, the resulting 3D points are sparse and noisy (see Figure~\ref{fig:mapping}(a)) which makes it difficult to use them for collision-free navigation. This is why we only use the refined poses to position and orient the 3D point clouds computed using the learned stereo module. The result is a dense and clean 3D geometric reconstruction (see Figure~\ref{fig:mapping}(b)).
\item \textbf{Mapping Store Items:} Our maps have to contain more than just a geometric reconstruction of the environment. They have to be enriched with the locations of grocery items in the store. To achieve this, we employ the object perception module described in Section~\ref{sec:object}. Since all camera poses are registered within a single coherent map coordinate system, we get the object location within the map by transforming its pointcloud according to the corresponding camera pose and computing the centroid. 
\end{itemize}

\noindent \textbf{Localization:} The robot uses the map generated offline to localize itself within the store. At the beginning of a shopping task execution, the robot estimates its pose relative to the map using the same procedure outlined in the above paragraph on loop detection and closure, where $L_k$ is the current image from the robot's chassis camera. Once localized, it relies on VO to navigate to each item in the shopping list. To correct for the VO drift, the robot re-localizes each time it arrives at an item.

\begin{figure}[t]
\includegraphics[width=\columnwidth]{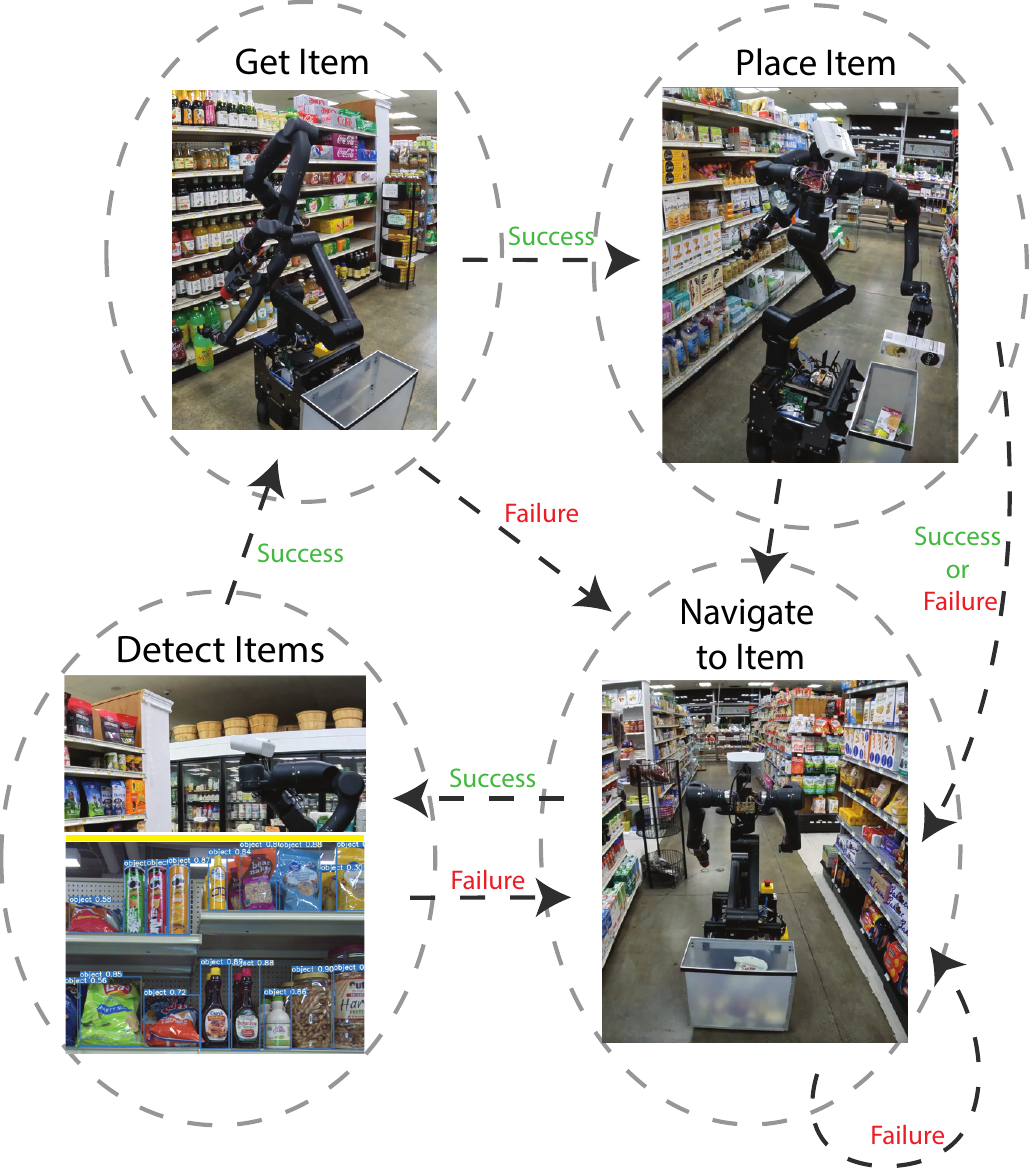}
\caption{Highly simplified task summary. Note that our hierarchical finite state machine includes many more states/transitions, but this provides a pictorial description of the main components of the grocery task.}
\label{fig:fsm}
\vspace{-0.5cm}
\end{figure}

\section{Planning} \label{sec:plan}
\subsection{Task Planning} \label{sec:plan_task}
At the highest level, our task planner is implemented as a hierarchical FSM, which guides the high level actions of the robot. 
For example, when the localization/navigation modules indicate that the robot is near the item to be grasped, a transition into a \textit{detection} state is triggered, where the robot runs object detection, segmentation, and classification (as detailed in Section~\ref{sec:object}) to find the requested item. A simplified illustration of the highest hierarchy of our FSM is shown in Figure~\ref{fig:fsm}. 

\subsection{Navigation} \label{sec:nav}
The first step of the navigation pipeline is to generate a 2.5D elevation map from the geometric reconstruction described in Section~\ref{sec:lm}. At the beginning of a shopping run, using that same geometric reconstruction with the (offline) mapped items, navigation goals (2D locations on the ground plane) are generated for each item in the randomly-generated shopping list by searching for a collision-free pose of a planar robot collision body along the outward-facing-axis of the item coordinate frame.  Obstacle-free paths (a sequence of 2D points) are then generated between all permutations of the items in the shopping list using an A$^*$ search within an inflated obstacle map generated from the 2.5D elevation map. The Christofides algorithm \cite{Christofides-1976-Tech} is then applied to a graph of these obstacle-free paths to compute the approximate shortest path that visits all items for the generated shopping list. This gives us a full path (set of 2D waypoints) that takes the robot to all the items in our shopping list. To track this path, a path follower algorithm \cite{helmick2006slip} is used to generate collision-free trajectories between consecutive waypoints.

The path follower uses high-rate, smooth pose estimates from combining wheel odometry (200 Hz) and visual odometry (5 Hz). Only at the end of each path (when it arrives at the item) does it re-localize within the map (as described in Section~\ref{sec:lm}). This is to prevent pose noise caused by the localization updates from affecting the performance of the path follower during execution.

\subsection{Motion Planning} \label{sec:plan_motion}
\begin{figure*}[ht]
\centering
\includegraphics[width=0.85\textwidth]{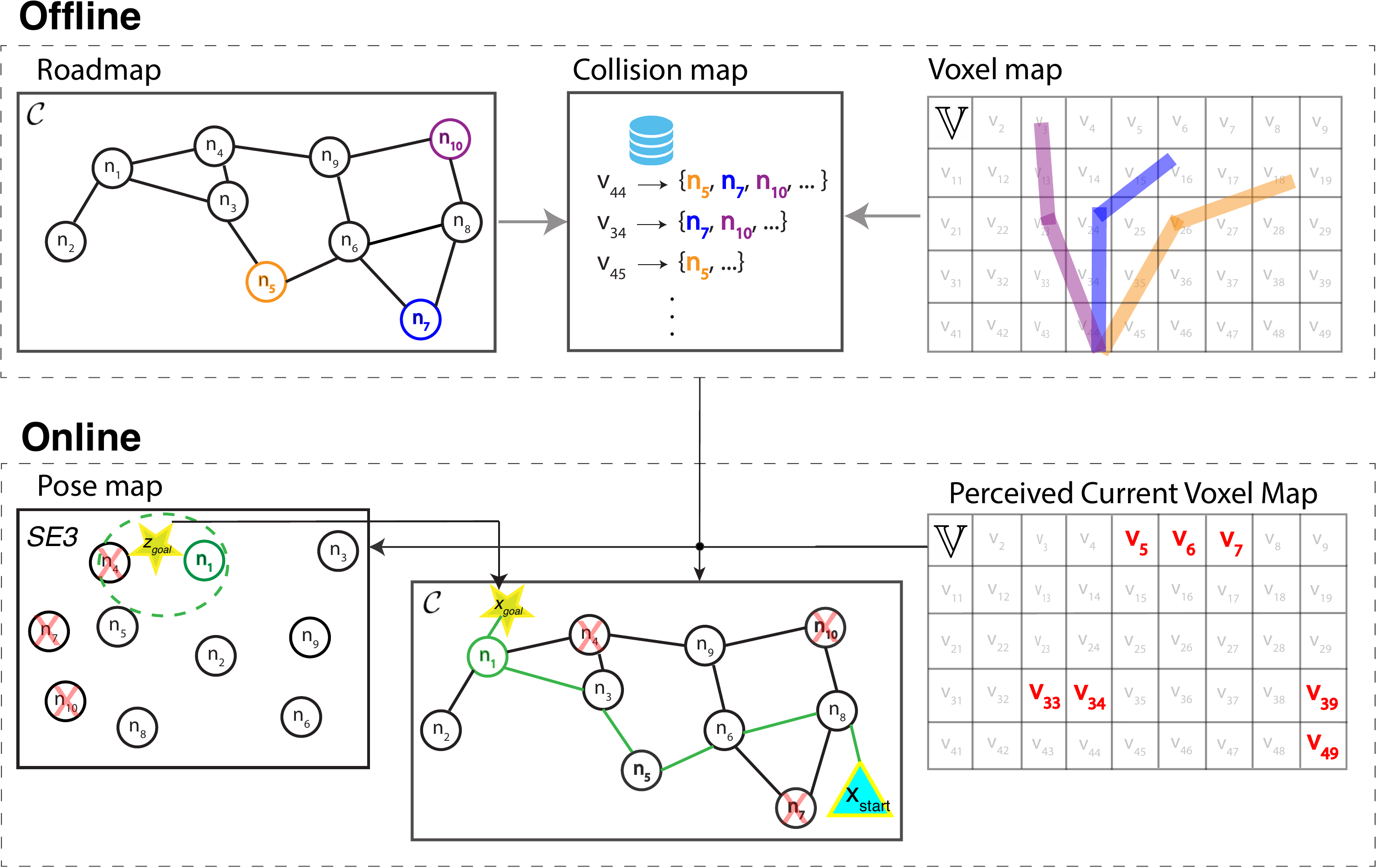}
\caption{Depiction of the motion planning pipeline. The top block denotes the offline process of building the roadmap and collision map of the DRM. The colored links in the voxel map (top right) depict the robot configurations corresponding to the same-colored nodes in the roadmap (top-left), such that the mapping from voxels to nodes-in-collision can be built. The bottom block denotes the online process of querying the DRM for a collision-free path, leveraging the collision map to avoid expensive collision checks.}
\label{fig:plan_motion}
\vspace{-0.4cm}
\end{figure*}
To accomplish different tasks, we must be able to command the robot to arbitrary target end-effector poses (e.g. for grasping objects and placing them) while avoiding collisions. For this purpose, we developed a motion planner that can achieve 100\% reliability with sub-second average planning times in a changing environment for our highly redundant, 21 DoF robot (the longest kinematic chain being 12 DoF). To describe the collision environment, we utilize the 3D voxel map mentioned in Section~\ref{sec:stereo}. Our robot leverages (1) a dynamic roadmap (DRM) \cite{Kallmann2004} that is pre-checked for collisions against any potential voxel map combined with (2) an optimization-based inverse kinematics (IK) solver \cite{Shankar2015} to quickly compute optimal plans in configuration space towards goals specified in Cartesian space. 

When a motion plan is requested, the voxel map is first processed and used to prune out all nodes in the DRM that are in collision. This is extremely fast, as these collision-checks are done offline based on the robot model and stored for efficient online queries. Since some collision-checks cannot be done offline (\eg, if the robot model changes because it grabbed an item), we leverage GPU-accelerated collision-checking for pruning out any nodes in collision with any newly added collision bodies. After this step, all remaining nodes in the DRM can be assumed to be collision-free, and a set of collision-free paths can be generated by searching for the shortest path to all nodes in the neighborhood of our target Cartesian pose (\eg, using A*). Finally, we rely on a QP-based IK solver to connect from nodes in this neighborhood of the target pose to the exact target pose, and the shortest resulting path is chosen. \footnote{A shortcutting-step is added at the end in order to optimize the robot path.}

The goals of this approach, outlined in Figure~\ref{fig:plan_motion}, are to front-load as much computation as possible to offline pre-computation (\ie, \textit{training} a large DRM) and to parallelize expensive aspects of online planning (\ie, collision-checks). 

\subsection{Grasp Planning} \label{sec:plan_grasp}
Upon successfully navigating to a target item's location as specified in the map and detecting the actual object instance in the live camera images, the robot plans how to pick this item off the grocery shelf and place it into its shopping basket. This requires the robot to take the following steps:
\begin{itemize}
\item Plan a collision-free path to the front of the desired item.
\item Reach into a cluttered shelf with either the parallel-jaw gripper or the suction gripper.
\item Grasp the item via a pinch or suction grasp.
\item Extract the item from the shelf.
\item Plan a collision-free path to place the item in its basket.
\item Move the gripper over the basket and release the item.
\end{itemize}
Before the robot extends its end-effector tool into the shelf, it corrects the pose of the tool, compensating for any kinematic errors, using a custom, GPU-accelerated point cloud registration method. After extraction, the robot evaluates whether the grasp was successful or not by examining the wrench at the tip and (a) the position of the gripper fingers or (b) pressure signals from the suction gripper.

\begin{figure*}[t]
\centering
\includegraphics[width=0.68\textwidth]{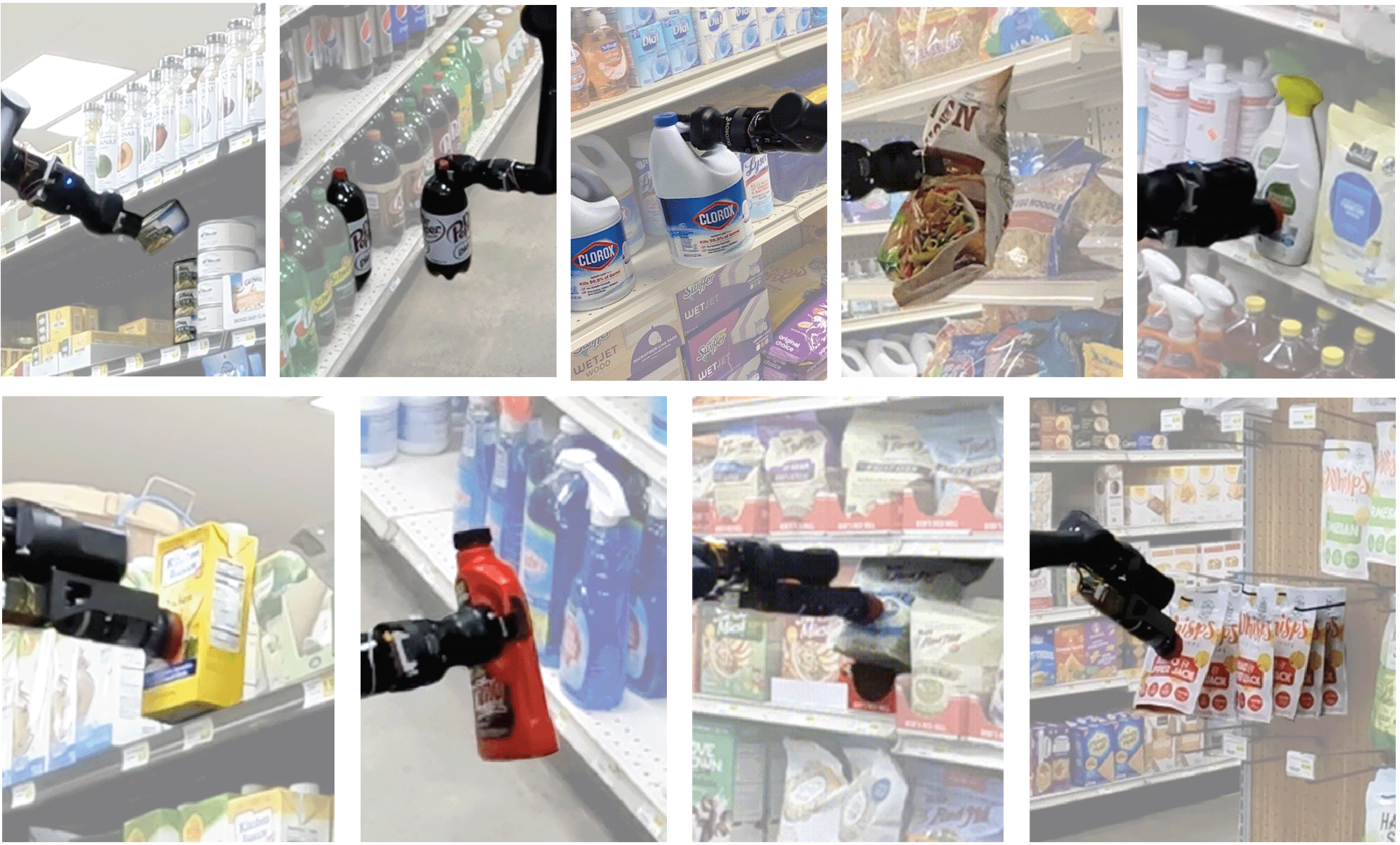}
\caption{Object Grasp Examples (top, left to right): (1.i) flat, cylindrical object grasp; (1.ii) cap grasp; (1.iii) handle grasp; (1.iv) heavy, deformable object grasp; (1.v) suction grasp.
(bottom, left to right): (2.i) extraction from a lip-free shelf; (2.ii) extraction of a jug; (2.iii) extraction from a box; (2.iv) extraction from a hook.}
\label{fig:grasp_1}
\vspace{-0.4cm}
\end{figure*}

\begin{figure}[t]
\includegraphics[width=\columnwidth]{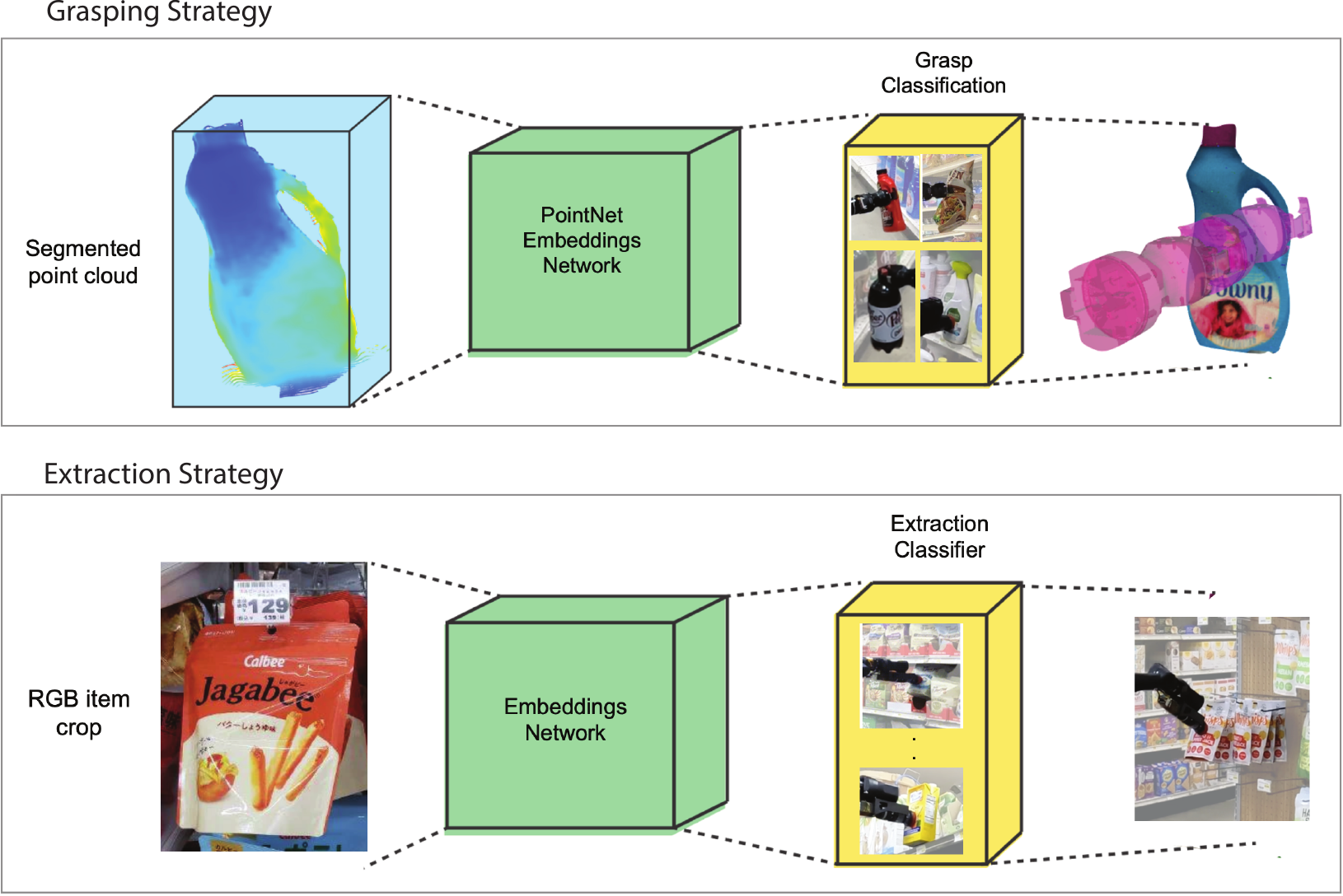}
\caption{\textbf{Top:} Network diagram for classifying grasp type based on segmented point cloud of item. This is combined with the pose of the object (computed downstream) to obtain a 6D grasp pose. \textbf{Bottom:} Network diagram for classifying extraction type based on RGB crop (e.g. extracting from hook vs. from box).}
\label{fig:classification_arichtecture}
\vspace{-0.6cm}
\end{figure}

In order to tackle grasping a variety of items in different environmental contexts (\eg, sitting on a shelf or hanging from a hook), we use a learned mapping from the perceived item to a discrete, comprehensive set of grasp/extract strategies for each item (ee Figure~\ref{fig:grasp_1}). Through experimentation with the robot, we found five different categories of grasps (\eg, grasp by handle, grasp by cap, suction at flattest region), and four categories of extraction (\eg, extract from hook) that handle most items represented in our chosen challenge task. We trained a PointNet-based \cite{PointNet} and a ResNet-based \cite{ResNet} network for grasp and extraction classification, respectively, with the grasp classifier using segmented point clouds and the extraction classifier using RGB crops of item instances as input. 
Once the type of grasp is inferred, a downstream process computes the grasp pose that is optimal for that grasp type. For example, for handle grasps, a keypoint detector was trained to find the center of handles in RGB images in order to anchor the grasp about that point. For convenience, classification of the grasp/extract type is done offline for every mapped item in the grocery store, while the grasp pose is computed online.
We use an engineered, geometry-based heuristic to select the object instance to pick if the robot detects multiple instances of a desired item type on the shelf. 

\section{Evaluation} \label{sec:eval}
\subsection{Task Description} \label{sec:task}
The challenge task requires the robot to fulfill a randomly generated shopping list in a real-world, unmodified\footnote{We install a mesh Wi-Fi network with Ethernet backhaul connections for the duration of our tests to ensure a stable connection between the operator station and the robot. However, this is not required since our robot is fully autonomous, and we do not consider it a modification of the environment.} grocery store. The items in the store are mapped prior to task execution using the procedure described in Section~\ref{sec:lm}. 
We conduct our tests at nighttime outside of the store's opening hours. This means no other shoppers are present and lighting conditions in the store are not affected by sunlight.

At the start of the task, the robot localizes itself relative to the map. Next, a shopping list of 20 unique items is generated randomly, where up to two instances of each item may be requested. The task is for the robot to autonomously collect all items in a basket and bring them to the starting point.

During the task, no human intervention is allowed, unless the robot is about to do something dangerous to itself or the environment. For those occasions, the human operator has a remote E-stop to end the task immediately. For liability and safety reasons, the robot operates outside of opening hours. Moreover, while no modifications are made to the grocery store, we narrow the scope of items we attempt to grasp primarily to avoid damaging products or the robot itself. Namely, we remove produce, items inside a refrigerator, items that are heavier than 4.5 kg\footnote{Note that TTT's arms are capable of handling up to 10 kg of payload, but most items in the grocery store that exceed 4.5 kg are too heavy for our suction tool and not graspable with a parallel-jaw gripper (e.g., heavy boxes of soda cans).}, and glass items. We will consider those in future work.

\subsection{Task Metrics} \label{sec:metrics}
Using the grocery shopping task as a way to continuously evaluate the performance of our system in an end-to-end fashion allows us to make data-driven decisions about where to focus our development efforts in order to efficiently improve the most important robot capabilities. To gather the required data, we conduct field tests every three months in an actual grocery store, during which the shopping task is executed as often as possible, over the course of five consecutive nights for four hours per night. We do not change the hardware or software of the robot system for the duration of this week-long test.

Based on detailed data logs, videos and records of all attempted picks, we use several metrics for performance assessment. In the following, we define the most important ones.

\noindent \textbf{Reliability:}
A task is considered completed if the robot navigates to all 20 items, attempts to pick the requested number of instances, and drives back to its starting point. Note that this definition does not take into account how many items were successfully picked and placed in the basket. Tasks are considered unsuccessful when they were prematurely stopped by an unrecoverable hardware or software fault, or by an operator-issued E-stop.

We define the \textit{task success rate} as the ratio of the number of completed tasks to the number of started tasks. We use this as our highest-level indicator of system reliability. Since any of the robot’s subsystems, from hardware to motion planning, can cause a task-ending failure, we use finer-grained indicators to give us more insight into what caused a specific failure. The unforgiving, multiplicative nature of the overall system reliability, measured by the task success rate, strongly motivates the development of fault recovery strategies for all subsystems.

\noindent \textbf{Shopping performance:} We measure the shopping performance as the ratio of the number of successfully retrieved items to the total number of items in the shopping list. Even though this metric can be negatively affected by items that are in the shopping list but out of stock, it gives us valuable insights into the system’s performance.

\noindent
\textbf{Speed:} We quantify speed as the overall time it takes the robot to complete one shopping list, divided by the number of items it retrieved. This overall speed metric incentivizes us to optimize speed on all levels, primarily for navigation, motion planning and motion execution.

\begin{figure}[t]
\includegraphics[width=\columnwidth]{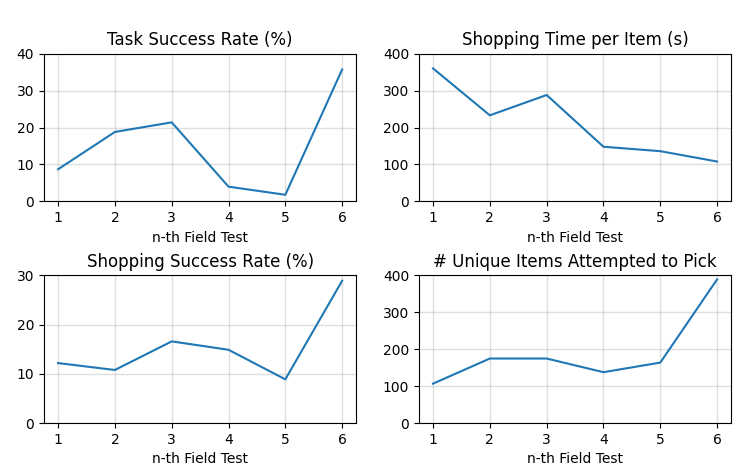}
\caption{Metrics that we use to drive our development. \textbf{Top Left:} Overall system reliability measured by the task success rate. \textbf{Top Right:} Speed measured by the shopping time per item. \textbf{Bottom Left:} Shopping performance measured by the shopping success rate. \textbf{Bottom Right:} Number of unique items attempted to grasp.}
\label{fig:results}
\vspace{-0.4cm}
\end{figure}

\subsection{Results} \label{sec:results}
Guided by the metrics described above, we have worked on improving our system’s performance in the grocery store shopping task for 18 months. In this time period, we conducted six field tests at a local grocery store at three months intervals. 
From the detailed log data collected during these tests, we track the above metrics that allow us to assess the overall performance and identify the most meaningful areas of improvement.

\begin{figure*}[t]
\centering
\includegraphics[width=0.9\textwidth]{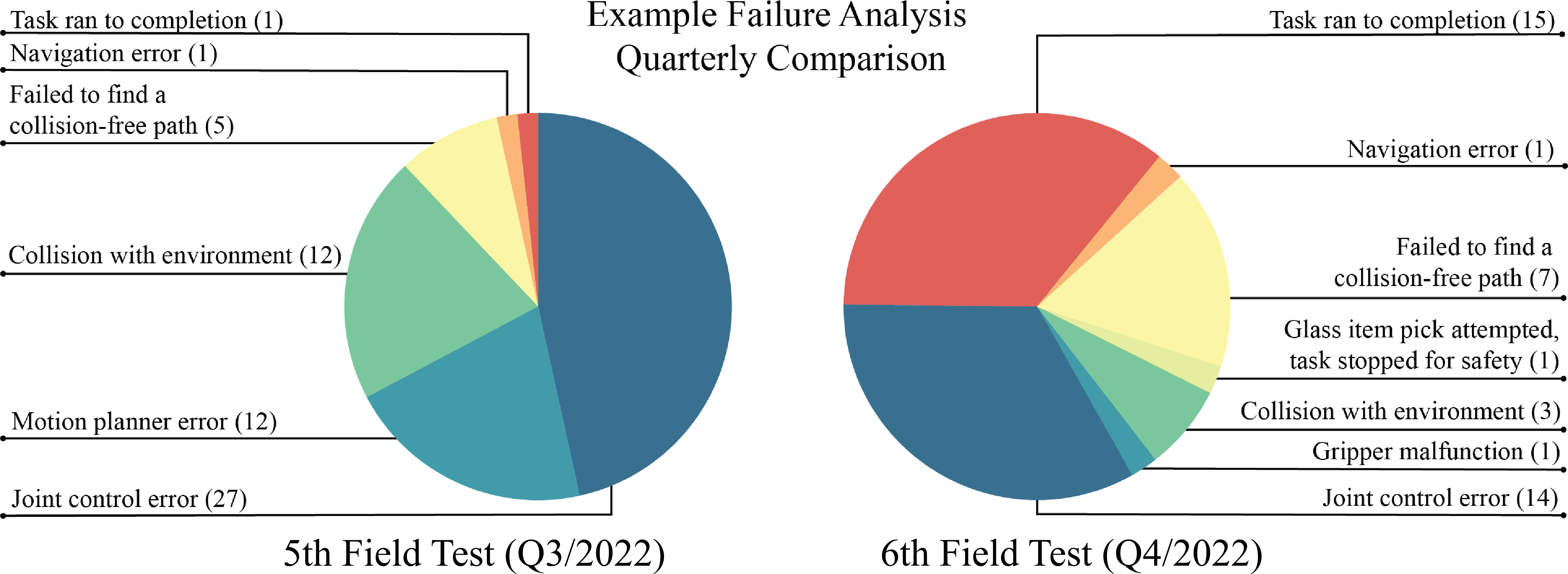}
\caption{Example of a comparison of quarterly field test failure analysis results. The pie charts break down causes of task failures, with the left and right charts corresponding to the fifth and the sixth field test, respectively.}
\label{fig:fa_comparison}
\vspace{-0.4cm}
\end{figure*}

Besides these high-level metrics, data from the field tests enables us to perform fine-grained analysis of all system failures. A visualization of such analyses for two consecutive quarters is shown in Figure~\ref{fig:fa_comparison}, where the main focus areas were reducing hardware and low-level software reliability problems with the new TTT platform ("joint control errors") and avoiding collisions with the environment.

\noindent \textbf{Reliability:}
Reliability, measured by the task success rate, is crucial for system up-time, a necessary requirement to collect the data for our data-driven development approach. 

We initially started this project with a different robot platform, FMT. While kinematically and morphologically almost identical, FMT is more heterogeneous in terms of hardware components, and its arms have a lower payload capacity than our current TTT platform. 
With FMT, we achieved a 21.4\% task success rate by the third field test. For the following field tests, we switched to the newly developed TTT platform. Naturally, this led to a decrease in system reliability as we worked through the challenges of bringing up a new platform. After the associated drop, system reliability recovered and got better than ever: the success rate reached 35.7\% by the sixth field test (see Figure~\ref{fig:results}, top left). 

This number is low compared to human performance, giving us lots of room to improve. However, we argue that it is high for a complex robot system operating in a real-world setting, considering that a shopping run can last up to 30 minutes and consists of hundreds of consecutive actions (such as planning, moving, driving, grasping). Even if each individual action was 99\% reliable, it only takes 103 consecutive actions for the task success rate to drop below the 35.7\% achieved by our robot.

\noindent \textbf{Shopping performance:}
We regard the shopping success rate (the ratio of the successfully retrieved items to the number of items in the shopping list) as the best overall indicator of shopping performance. 
It is sensitive to almost every aspect of the system’s performance, except for speed. 
Largely driven by the significant increase in our system’s reliability between the last two field tests, the overall performance has started a very exciting upward trend from 8.9\% in the fifth to 28.9\% in the sixth field test (see Figure~\ref{fig:results}, bottom left). 
This increased performance and the resulting higher number of grasp executions per field test unlock much more effective data collection, which will further accelerate our development.

\noindent \textbf{Speed:}
We made continuous overall progress on the shopping speed, which is measured by the \textit{shopping time per item}. 
This time includes navigating to an item as well as picking it. 
We were able to reduce this metric by approximately 70\% over the course of six field tests, from six minutes per item to below two minutes (see Figure~\ref{fig:results}, top right). 
The largest increase in speed occurred when we switched from the FMT to the more capable TTT robot platform. 
However, switching to a faster robot was only one aspect of increasing the speed. 
More importantly, we developed tools to investigate which parts of the shopping task take up significant time and were able to focus our speed-up efforts there. 
This led to progress in widely applicable robot capabilities such as faster motion planning for redundant systems, faster collision-free navigation in confined spaces, and faster grasping.

\noindent \textbf{Unique items attempted to pick:}
Of the many more detailed metrics we track, one that is particularly intuitive and valuable to us is the \textit{number of unique items attempted to pick} during a field test.
We aim to steadily increase this number, which makes the overall problem harder, without regressing on the other metrics.
It is affected by our mapping capabilities and the robot’s ability to grasp a wide variety of items with its two complementary tools, but also by the sheer time the robot is actually executing shopping runs, which is in turn affected by its reliability. 
Thanks to a map containing close to 1000 items, an ever more capable grasping pipeline and the significant uptick in system reliability, this metric has entered a steep upward trend from 164 unique items in the fifth field test to 389 in the sixth (see Figure~\ref{fig:results}, bottom right). 
Each attempted pick, successful or not, creates invaluable data from which we can learn and improve. 
For this reason we expect this increase to benefit many aspects of our development pipeline.

\section{Lessons Learned and Key Takeaways} \label{sec:key}
\label{sec:conclusion}
Based on our extensive experiments in the grocery store and analysis of the resulting data (including successes and failures), we have compiled key takeaways that we believe could accelerate deployment of robots in the wild.

\noindent \textbf{System/component reliability is the most important metric for real-world deployment:}
\begin{itemize}
\item Working on end-to-end systems with many actions chained in series emphasizes how important reliability and accuracy of the individual components/subsystems are. If each individual action is 99\% reliable, and we take 10 actions per minute, overall reliability would be 0.2\% for a task that takes 1 hour. 
\item Resetting the robot is not a viable option in the real-world. When the robot gets stuck or hits an object or goes down unexpectedly, human intervention is required, at least partially defeating the purpose of an autonomous robot.
\item Homogeneous hardware enables more reliable systems. TTT, using our in-house actuators in all of its joints, is more reliable than our previous, more heterogeneous robot architectures. Our observation is that more homogeneous systems require fewer unique hardware and software components. These components are therefore easier to test, improve and harden. 
\item In order to see research results transition to real-world applications, there needs to be a viable path from proof-of-concept towards making methods reliable. While ‘hardening’ certain methods is often deemed an engineering rather than a research problem, we want to challenge this notion. Making methods that are powerful and promising in a laboratory setting more robust or fault-tolerant to real-world challenges can be seen as a research problem in its own right, requiring creative ideas and sometimes entirely new approaches. 
Switching robot platforms in the middle of the project highlighted the need for methods that generalize beyond a highly specific hardware or system architecture.
\end{itemize}

\noindent \textbf{Machine learning has revolutionized many aspects of robotics and enabled several new capabilities, but we are still far from truly intelligent robots:}

\begin{itemize}
\item While we use learning methods for narrow tasks such as stereo reconstruction and gripper selection, a \textit{reliable} learning-based method to synthesize broader robot behaviors (including fault responses) is still missing, but it is exciting to see recent progress towards this vision (\eg, \cite{rt12022arxiv}).
\item Explainable AI will be crucial for deployment of powerful learned models in mobile manipulation. The lack of interpretability of current learned methods make fault recovery and failure analysis extremely difficult, with the only choice often being to add more/better data. Therefore, we are very excited to see advancements in this area leading to more reliable deployment in mobile manipulation robots.
\end{itemize}

\noindent \textbf{Manipulation with a mobile robot requires constant hardware-software co-design and co-evolution:}
\begin{itemize}
\item A robot that can navigate and manipulate objects within a human-centric environment benefits from a smaller footprint. In a sense, this places hardware and software development at tension with each other – smaller actuators are more difficult to design and manufacture, yet larger robots face more challenging collision constraints, affecting all areas of planning. 
\item Constant trade-offs are necessary when considering end-effector designs for mobile manipulators, including balancing safety vs. strength, simplicity vs. versatility, and power vs. form-factor. For instance, grasps for suction grippers are easier to generate compared to grasps for mechanical grippers. However, designing suction grippers that can successfully grasp a wide variety of materials and weights while maintaining a small form-factor poses new challenges in hardware design.
\end{itemize}

\section{Discussion}

We hope we have impressed upon the reader that mobile manipulation is a hard problem, with the need and many exciting opportunities for future research. 
Two things we deem critical for continued progress are: (1) end-to-end system testing, and (2) field testing in real-world (non-lab) environments. 

End-to-end system testing (in addition to unit tests and integration tests) is necessary because algorithms as well as hardware subsystems that perform well in isolation may not perform well on a complex robotic system. 
Concrete examples that we encountered include: 
\begin{itemize}
\item Our voxel mapper reconstructs and tracks obstacles. It uses the robot kinematics to distinguish the robot itself from the environment by masking out robot pixels, preventing them from entering the voxel mapper pipeline. 
Our motion planner relies on an accurate voxel map for collision free planning. While both these components functioned well (passing all unit tests), end-to-end testing revealed rare instances in which kinematic uncertainty leads to parts of the robot erroneously being labeled as environment, preventing successful motion planning. This led us to (1) develop a purely vision-based robot mask, independent from the robot kinematics, and (2) make our motion planner robust to spurious voxels arising from an imperfect robot mask.
\item If an item to grasp is far enough from the current robot position, requiring our gripper to make a difficult reach, the robot first does a lateral drive to make the grasp easier. 
To keep the item in view, the robot needs to rotate its head, resulting in skewed images. We learned that this leads to a much higher likelihood of failed or inaccurate item perception. This observation led us to re-think this aspect of the overall item retrieval pipeline.
\end{itemize}
There are many more examples of such subtle interplay that are difficult to produce when testing modules in isolation, but that arise during long periods of end-to-end testing. 
Such instances are important for highlighting inaccurate assumptions made by different algorithms so that we can push towards addressing them.

In addition to end-to-end testing, periodic testing in real-world environments is crucial as it will highlight issues that do not arise in the lab, and expose implicit assumptions that hold true in the lab but not in the field. 
As an analogy, consider how in the machine learning field, it is standard practice to divide a dataset into a training set, validation set, and test set. 
The validation set can be used to tune hyperparameters of the learning algorithm, whereas the test set should only ever be used for evaluation. 
The reasoning is that having access to the validation set, even just to tune hyperparameters, introduces the risk of overfitting to it.
\textit{A lab environment is analogous to a validation set} - even if it is designed to approximate the real world, we risk overfitting to its specific characteristics.
\textit{Testing in the field is analogous to testing on a test set}.

We have found that our metrics are significantly better in the lab (i.e. our mock grocery store) versus the real-world grocery store. 
Part of that is because the real environment is more challenging (e.g. customers place items in odd configurations, the store layout and placement of items is constantly changing), but a large part is also that we have constant access to the lab when designing/tuning our algorithms, and so we can unconsciously overfit to the challenges we encounter there.

\section*{Acknowledgments}

We would like to thank JC Hancock, Jordan Skerbetz, Andrew Custer and Jonathan Yao for their support with map data collection and robot testing.

\bibliographystyle{plainnat}
\bibliography{references}

\begin{thebibliography}{37}
\providecommand{\natexlab}[1]{#1}
\providecommand{\url}[1]{\texttt{#1}}
\expandafter\ifx\csname urlstyle\endcsname\relax
  \providecommand{\doi}[1]{doi: #1}\else
  \providecommand{\doi}{doi: \begingroup \urlstyle{rm}\Url}\fi

\bibitem[Agarwal et~al.(2022)Agarwal, Mierle, and Team]{Agarwal-2022}
Sameer Agarwal, Keir Mierle, and The Ceres~Solver Team.
\newblock {Ceres Solver}, 2022.
\newblock URL \url{https://github.com/ceres-solver/ceres-solver}.

\bibitem[Arandjelovic(2012)]{Arandjelovic-2012-CVPR}
Relja Arandjelovic.
\newblock {Three Things Everyone Should Know to Improve Object Retrieval}.
\newblock In \emph{Proceedings of the IEEE Conference on Computer Vision and
  Pattern Recognition (CVPR)}, 2012.

\bibitem[Arandjelovic et~al.(2016)Arandjelovic, Gronat, Torii, Pajdla, and
  Sivic]{Arandjelovic-2016-CVPR}
Relja Arandjelovic, Petr Gronat, Akihiko Torii, Tomas Pajdla, and Josef Sivic.
\newblock {NetVLAD: CNN Architecture for Weakly Supervised Place Recognition}.
\newblock In \emph{Proceedings of the IEEE Conference on Computer Vision and
  Pattern Recognition (CVPR)}, 2016.

\bibitem[Asfour et~al.(2019)Asfour, Waechter, Kaul, Rader, Weiner, Ottenhaus,
  Grimm, Zhou, Grotz, and Paus]{asfour-2019-armar}
Tamim Asfour, Mirko Waechter, Lukas Kaul, Samuel Rader, Pascal Weiner, Simon
  Ottenhaus, Raphael Grimm, You Zhou, Markus Grotz, and Fabian Paus.
\newblock {ARMAR-6: A High-Performance Humanoid for Human-Robot Collaboration
  in Real-World Scenarios}.
\newblock \emph{IEEE Robotics \& Automation Magazine}, 26\penalty0 (4), 2019.

\bibitem[Bajracharya et~al.(2012)Bajracharya, Ma, Howard, and
  Matthies]{bajracharya2012real}
Max Bajracharya, Jeremy Ma, Andrew Howard, and Larry Matthies.
\newblock {Real-Time 3D Stereo Mapping in Complex Dynamic Environments}.
\newblock In \emph{Proceedings of the International Conference on Robotics and
  Automation-Semantic Mapping, Perception, and Exploration (SPME) Workshop},
  volume~15, 2012.

\bibitem[Bajracharya et~al.(2020)Bajracharya, Borders, Helmick, Kollar, Laskey,
  Leichty, Ma, Nagarajan, Ochiai, Petersen, Stone, and
  Takaoka]{bajracharya2020mobile}
Max Bajracharya, James Borders, Dan Helmick, Thomas Kollar, Michael Laskey,
  John Leichty, Jeremy Ma, Umashankar Nagarajan, Akiyoshi Ochiai, Josh
  Petersen, Kevin Stone, and Yutaka Takaoka.
\newblock {A Mobile Manipulation System for One-Shot Teaching of Complex Tasks
  in Homes}.
\newblock In \emph{Proceedings of the IEEE International Conference on Robotics
  and Automation (ICRA)}, 2020.

\bibitem[Borst et~al.(2009)Borst, Wimbock, Schmidt, Fuchs, Brunner, Zacharias,
  Giordano, Konietschke, Sepp, and Fuchs]{borst-2009-rollin}
Christoph Borst, Thomas Wimbock, Florian Schmidt, Matthias Fuchs, Bernhard
  Brunner, Franziska Zacharias, Paolo~Robuffo Giordano, Rainer Konietschke,
  Wolfgang Sepp, and Stefan Fuchs.
\newblock {Rollin'Justin-Mobile Platform with Variable Base}.
\newblock In \emph{Proceedings of the IEEE International Conference on Robotics
  and Automation (ICRA)}, 2009.

\bibitem[Brohan et~al.(2022)Brohan, Brown, Carbajal, Chebotar, Dabis, Finn,
  Gopalakrishnan, Hausman, Herzog, Hsu, Ibarz, Ichter, Irpan, Jackson,
  Jesmonth, Joshi, Julian, Kalashnikov, Kuang, Leal, Lee, Levine, Lu, Malla,
  Manjunath, Mordatch, Nachum, Parada, Peralta, Perez, Pertsch, Quiambao, Rao,
  Ryoo, Salazar, Sanketi, Sayed, Singh, Sontakke, Stone, Tan, Tran, Vanhoucke,
  Vega, Vuong, Xia, Xiao, Xu, Xu, Yu, and Zitkovich]{rt12022arxiv}
Anthony Brohan, Noah Brown, Justice Carbajal, Yevgen Chebotar, Joseph Dabis,
  Chelsea Finn, Keerthana Gopalakrishnan, Karol Hausman, Alex Herzog, Jasmine
  Hsu, Julian Ibarz, Brian Ichter, Alex Irpan, Tomas Jackson, Sally Jesmonth,
  Nikhil Joshi, Ryan Julian, Dmitry Kalashnikov, Yuheng Kuang, Isabel Leal,
  Kuang-Huei Lee, Sergey Levine, Yao Lu, Utsav Malla, Deeksha Manjunath, Igor
  Mordatch, Ofir Nachum, Carolina Parada, Jodilyn Peralta, Emily Perez, Karl
  Pertsch, Jornell Quiambao, Kanishka Rao, Michael Ryoo, Grecia Salazar, Pannag
  Sanketi, Kevin Sayed, Jaspiar Singh, Sumedh Sontakke, Austin Stone, Clayton
  Tan, Huong Tran, Vincent Vanhoucke, Steve Vega, Quan Vuong, Fei Xia, Ted
  Xiao, Peng Xu, Sichun Xu, Tianhe Yu, and Brianna Zitkovich.
\newblock {RT-1: Robotics Transformer for Real-World Control at Scale}.
\newblock In \emph{arXiv preprint arXiv:2212.06817}, 2022.

\bibitem[Christofides(1976)]{Christofides-1976-Tech}
Nicos Christofides.
\newblock {Worst-Case Analysis of a New Heuristic for the Travelling Salesman
  Problem}.
\newblock Technical Report 388, Graduate School of Industrial Administration,
  Carnegie Mellon University, 1976.

\bibitem[D{\"o}mel et~al.(2017)D{\"o}mel, Kriegel, Ka{\ss}ecker, Brucker,
  Bodenm{\"u}ller, and Suppa]{domel-2017-toward}
Andreas D{\"o}mel, Simon Kriegel, Michael Ka{\ss}ecker, Manuel Brucker, Tim
  Bodenm{\"u}ller, and Michael Suppa.
\newblock {Toward Fully Autonomous Mobile Manipulation for Industrial
  Environments}.
\newblock \emph{International Journal of Advanced Robotic Systems}, 14\penalty0
  (4), 2017.

\bibitem[Eppner et~al.(2016)Eppner, H{\"o}fer, Jonschkowski,
  Mart{\'\i}n-Mart{\'\i}n, Sieverling, Wall, and Brock]{eppner2016lessons}
Clemens Eppner, Sebastian H{\"o}fer, Rico Jonschkowski, Roberto
  Mart{\'\i}n-Mart{\'\i}n, Arne Sieverling, Vincent Wall, and Oliver Brock.
\newblock Lessons from the {A}mazon picking challenge: Four aspects of building
  robotic systems.
\newblock In \emph{Robotics: science and systems}, pages 4831--4835, 2016.

\bibitem[Fischler and Bolles(1981)]{Fischler-1981-ACM}
Martin~A. Fischler and Robert~C. Bolles.
\newblock {Random Sample Consensus: A Paradigm for Model Fitting with
  Applications to Image Analysis and Automated Cartography}.
\newblock \emph{Communications of the ACM}, 24\penalty0 (6), 1981.

\bibitem[Glenn(2020)]{Glenn-2020}
Jocher Glenn.
\newblock {YOLOv5 by Ultralytics}, 2020.
\newblock URL \url{https://github.com/ultralytics/yolov5}.

\bibitem[Goldman et~al.(2019)Goldman, Herzig, Eisenschtat, Goldberger, and
  Hassner]{Goldman-2019-CVPR}
Eran Goldman, Roei Herzig, Aviv Eisenschtat, Jacob Goldberger, and Tal Hassner.
\newblock {Precise Detection in Densely Packed Scenes}.
\newblock In \emph{Proceedings of the IEEE/CVF Conference on Computer Vision
  and Pattern Recognition (CVPR)}, 2019.

\bibitem[Guizzo and Ackerman(2015)]{Guizzo-2015-Spectrum}
Erico Guizzo and Evan Ackerman.
\newblock {The Hard Lessons of DARPA's Robotics Rhallenge [News]}.
\newblock \emph{IEEE Spectrum}, 52\penalty0 (8), 2015.

\bibitem[He et~al.(2016)He, Zhang, Ren, and Sun]{ResNet}
Kaiming He, Xiangyu Zhang, Shaoqing Ren, and Jian Sun.
\newblock {Deep Residual Learning for Image Recognition}.
\newblock In \emph{Proceedings of the IEEE Conference on Computer Vision and
  Pattern Recognition (CVPR)}, 2016.

\bibitem[Hebert et~al.(2015)Hebert, Bajracharya, Ma, Hudson, Aydemir, Reid,
  Bergh, Borders, Frost, Hagman, Leichty, Backes, Kennedy, Karplus, Satzinger,
  Byl, Shankar, and Burdick]{robosimian}
Paul Hebert, Max Bajracharya, Jeremy Ma, Nicolas Hudson, Alper Aydemir, Jason
  Reid, Charles Bergh, James Borders, Matthew Frost, Michael Hagman, John
  Leichty, Paul Backes, Brett Kennedy, Paul Karplus, Brian Satzinger, Katie
  Byl, Krishna Shankar, and Joel Burdick.
\newblock {Mobile Manipulation and Mobility as Manipulation—Design and
  Algorithms of RoboSimian}.
\newblock \emph{Journal of Field Robotics}, 32\penalty0 (2), 2015.

\bibitem[Helmick et~al.(2006)Helmick, Roumeliotis, Cheng, Clouse, Bajracharya,
  and Matthies]{helmick2006slip}
Daniel~M Helmick, Stergios~I Roumeliotis, Yang Cheng, Daniel~S Clouse, Max
  Bajracharya, and Larry~H Matthies.
\newblock {Slip-Compensated Path Following for Planetary Exploration Rovers}.
\newblock \emph{Advanced Robotics}, 20\penalty0 (11), 2006.

\bibitem[Kallmann and Mataric(2004)]{Kallmann2004}
Marcelo Kallmann and Maja Mataric.
\newblock {Motion Planning using Dynamic Roadmaps}.
\newblock In \emph{Proceedings of the IEEE International Conference on Robotics
  and Automation (ICRA)}, 2004.

\bibitem[Koch et~al.(2015)Koch, Zemel, and Salakhutdinov]{Koch-2015-ICML}
Gregory Koch, Richard Zemel, and Ruslan Salakhutdinov.
\newblock {Siamese Neural Networks for One-Shot Image Recognition}.
\newblock In \emph{ICML deep learning workshop}, volume~2, 2015.

\bibitem[Kraetzschmar et~al.(2015)Kraetzschmar, Hochgeschwender, Nowak, Hegger,
  Schneider, Dwiputra, Berghofer, and Bischoff]{Kraetzschmar-2015-Robo}
Gerhard~K. Kraetzschmar, Nico Hochgeschwender, Walter Nowak, Frederik Hegger,
  Sven Schneider, Rhama Dwiputra, Jakob Berghofer, and Rainer Bischoff.
\newblock {RoboCup@Work: Competing for the Factory of the Future}.
\newblock In \emph{RoboCup 2014: Robot World Cup XVIII}, 2015.

\bibitem[Lepetit et~al.(2009)Lepetit, Moreno-Noguer, and
  Fua]{Lepetit-2009-IJCV}
Vincent Lepetit, Francesc Moreno-Noguer, and Pascal Fua.
\newblock {EPnP: An Accurate O(n) Solution to the PnP Problem}.
\newblock \emph{International Journal Computer Vision}, 81\penalty0 (2), 2009.

\bibitem[Lindeberg(1993)]{Lindeberg-1993-IJCV}
Tony Lindeberg.
\newblock {Detecting Salient Blob-Like Image Structures and Their Scales with a
  Scale-Space Primal Sketch: A method for Focus-of-Attention}.
\newblock \emph{International Journal Computer Vision}, 11\penalty0 (3), 1993.

\bibitem[Liu et~al.(2020)Liu, Kang, Li, Hua, and Vasconcelos]{Liu-2020-CVPR}
Bo~Liu, Hao Kang, Haoxiang Li, Gang Hua, and Nuno Vasconcelos.
\newblock {Few-Shot Open-Set Recognition Using Meta-Learning}.
\newblock In \emph{Proceedings of the IEEE/CVF Conference on Computer Vision
  and Pattern Recognition (CVPR)}, June 2020.

\bibitem[Meeussen et~al.(2010)Meeussen, Wise, Glaser, Chitta, McGann, Mihelich,
  Marder-Eppstein, Muja, Eruhimov, Foote, Hsu, Rusu, Marthi, Bradski, Konolige,
  Gerkey, and Berger]{Meeussen-2010-ICRA}
Wim Meeussen, Melonee Wise, Stuart Glaser, Sachin Chitta, Conor McGann, Patrick
  Mihelich, Eitan Marder-Eppstein, Marius Muja, Victor Eruhimov, Tully Foote,
  John Hsu, Radu~Bogdan Rusu, Bhaskara Marthi, Gary Bradski, Kurt Konolige,
  Brian Gerkey, and Eric Berger.
\newblock {Autonomous Door Opening and Plugging in with a Personal Robot}.
\newblock In \emph{Proceedings of the IEEE International Conference on Robotics
  and Automation (ICRA)}, 2010.

\bibitem[Qi et~al.(2017)Qi, Su, Mo, and Guibas]{PointNet}
Charles~R Qi, Hao Su, Kaichun Mo, and Leonidas~J Guibas.
\newblock {Pointnet: Deep Learning on Point Sets for 3D Classification and
  Segmentation}.
\newblock In \emph{Proceedings of the IEEE Conference on Computer Vision and
  Pattern Recognition (CVPR)}, 2017.

\bibitem[Raghavan et~al.(2022)Raghavan, Kanoulas, Caldwell, and
  Tsagarakis]{centauro}
Vignesh~Sushrutha Raghavan, Dimitrios Kanoulas, Darwin~G. Caldwell, and
  Nikos~G. Tsagarakis.
\newblock {Reconfigurable and Agile Legged-Wheeled Robot Navigation in
  Cluttered Environments With Movable Obstacles}.
\newblock \emph{IEEE Access}, 10, 2022.

\bibitem[Ronneberger et~al.(2015)Ronneberger, Fischer, and Brox]{UNet}
Olaf Ronneberger, Philipp Fischer, and Thomas Brox.
\newblock {U-net: Convolutional Networks for Biomedical Image Segmentation}.
\newblock In \emph{Proceedings of the 18th International Conference on Medical
  Image Computing and Computer-Assisted Intervention (MICCAI)}, 2015.

\bibitem[Sereinig et~al.(2020)Sereinig, Werth, and Faller]{Sereinig-2020}
Martin Sereinig, Wolfgang Werth, and Lisa-Marie Faller.
\newblock {A Review of the Challenges in Mobile Manipulation: Systems Design
  and RoboCup Challenges}.
\newblock \emph{e {\&} i Elektrotechnik und Informationstechnik}, 137\penalty0
  (6), Oct 2020.

\bibitem[Shankar et~al.(2015)Shankar, Burdick, and Hudson]{Shankar2015}
Krishna Shankar, Joel~W. Burdick, and Nicolas~H. Hudson.
\newblock {A Quadratic Programming Approach to Quasi-Static Whole-Body
  Manipulation}.
\newblock In \emph{Algorithmic Foundations of Robotics XI: Selected
  Contributions of the Eleventh International Workshop on the Algorithmic
  Foundations of Robotics}, 2015.

\bibitem[Shankar et~al.(2022)Shankar, Tjersland, Ma, Stone, and
  Bajracharya]{Shankar-2022-RAL}
Krishna Shankar, Mark Tjersland, Jeremy Ma, Kevin Stone, and Max Bajracharya.
\newblock {A Learned Stereo Depth System for Robotic Manipulation in Homes}.
\newblock \emph{IEEE Robotics and Automation Letters}, 7\penalty0 (2), 2022.

\bibitem[Snell et~al.(2017)Snell, Swersky, and Zemel]{Snell-NIPS-2017}
Jake Snell, Kevin Swersky, and Richard Zemel.
\newblock {Prototypical Networks for Few-shot Learning}.
\newblock In \emph{Advances in Neural Information Processing Systems (NIPS)},
  volume~30, 2017.

\bibitem[Srinivasa et~al.(2012)Srinivasa, Berenson, Cakmak, Collet, Dogar,
  Dragan, Knepper, Niemueller, Strabala, Vande~Weghe, and
  Ziegler]{herb2_journal}
Siddhartha~S. Srinivasa, Dmitry Berenson, Maya Cakmak, Alvaro Collet, Mehmet~R.
  Dogar, Anca~D. Dragan, Ross~A. Knepper, Tim Niemueller, Kyle Strabala, Mike
  Vande~Weghe, and Julius Ziegler.
\newblock {Herb 2.0: Lessons Learned From Developing a Mobile Manipulator for
  the Home}.
\newblock \emph{Proceedings of the IEEE}, 100\penalty0 (8), 2012.

\bibitem[{\v{S}}tibinger et~al.(2021){\v{S}}tibinger, Broughton, Majer,
  Rozsyp{\'a}lek, Wang, Jindal, Zhou, Thakur, Loianno, Krajn{\'\i}k, and
  Saska]{vstibinger2021mobile}
Petr {\v{S}}tibinger, George Broughton, Filip Majer, Zden{\v{e}}k
  Rozsyp{\'a}lek, Anthony Wang, Kshitij Jindal, Alex Zhou, Dinesh Thakur,
  Giuseppe Loianno, Tom{\'a}{\v{s}} Krajn{\'\i}k, and Martin Saska.
\newblock {Mobile Manipulator for Autonomous Localization, Grasping and Precise
  Placement of Construction Material in a Semi-Structured Environment}.
\newblock \emph{IEEE Robotics and Automation Letters}, 6\penalty0 (2), 2021.

\bibitem[Thakar et~al.(2023)Thakar, Srinivasan, Al-Hussaini, Bhatt, Rajendran,
  Jung~Yoon, Dhanaraj, Malhan, Schmid, Krovi, and Gupta]{Thakar-2023-MaR}
Shantanu Thakar, Srivatsan Srinivasan, Sarah Al-Hussaini, Prahar~M Bhatt,
  Pradeep Rajendran, Yeo Jung~Yoon, Neel Dhanaraj, Rishi~K Malhan, Matthias
  Schmid, Venkat~N Krovi, and Satyandra~K. Gupta.
\newblock {A Survey of Wheeled Mobile Manipulation: A Decision-Making
  Perspective}.
\newblock \emph{Journal of Mechanisms and Robotics}, 15\penalty0 (2), 2023.

\bibitem[Whelan et~al.(2015)Whelan, Leutenegger, Salas-Moreno, Glocker, and
  Davison]{ElasticFusion}
Thomas Whelan, Stefan Leutenegger, Renato Salas-Moreno, Ben Glocker, and Andrew
  Davison.
\newblock {ElasticFusion: Dense SLAM without a Pose Graph}.
\newblock In \emph{Proceedings of Robotis Science and Systems (RSS)}, 2015.

\bibitem[Wisspeintner et~al.(2009)Wisspeintner, Van Der~Zant, Iocchi, and
  Schiffer]{wisspeintner-2009-robocup}
Thomas Wisspeintner, Tijn Van Der~Zant, Luca Iocchi, and Stefan Schiffer.
\newblock {RoboCup@ Home: Scientific Competition and Benchmarking for Domestic
  Service Robots}.
\newblock \emph{Interaction Studies}, 10\penalty0 (3), 2009.

\end{thebibliography}

\end{document}